\newcommand{\theabstract}{Early identification of cerebral palsy (CP) relies on observational assessments of spontaneous infant movement, such as the General Movement Assessment (GMA) and the Baby Observational Selective Control AppRaisal (BabyOSCAR), which require trained raters and substantial scoring time. Existing automated approaches derive hand-engineered features from two-dimensional keypoints, which describe image-plane motion rather than the three-dimensional behavior of individual joints. We present Baby Biomechanical 3D Body (BB3D), an infant-adapted, biomechanical pipeline that estimates joint-angle trajectories from single-camera video. Against a multi-view markerless motion capture reference in 17 infants, BB3D's median Procrustes-aligned error was 17.5~mm for joint centers and 20.4~mm for marker sites. A spatio-temporal transformer, pretrained on BB3D joint-angle trajectories from 1k+ single-camera recordings, and fine-tuned on 550 videos from 350 infants, predicted each individual BabyOSCAR item with a mean within-site AUC of 0.79 and the BabyOSCAR total with a Spearman correlation of 0.86 on held-out recordings. Without being trained on diagnosis, the predicted total separated infants with CP from those without CP with an AUC of 0.86, exceeding a replicated two-dimensional feature classifier (0.67) but below the expert clinician rater's total (0.97). Predicted scores also carried information about functional ability and body distribution, though less than expert clinician scores. These results show that biomechanically grounded representations of monocular video can support scalable, joint-level assessment of selective motor control in infancy.}

\documentclass{isrpaper}

\usepackage{tabularx}
\newcolumntype{C}{>{\centering\arraybackslash}X}
\usepackage{makecell}
\usepackage{threeparttable}
\usepackage{graphicx}
\usepackage{float}
\usepackage{amsmath,amssymb,amsfonts}
\usepackage{textcomp}
\usepackage{xcolor}
\usepackage{booktabs}
\usepackage{ragged2e}  % allows \justifying

\usepackage{orcidlink}

\usepackage[
    backend=biber,
    style=authoryear,
    dashed=false,
    natbib=true,
    doi=false,
    url=false,
    isbn=false,
    maxbibnames=99,
    maxcitenames=2,
    uniquelist=false,
    giveninits=true,    % "Cotton, R. J." instead of "Cotton, R. James"
    terseinits=true,    % "R.J." instead of "R. J."
]{biblatex}

\AtEveryBibitem{%
   \clearfield{Title}%
   \clearfield{eprint}%
   \clearfield{note}%
}

\renewcommand{\cite}[1]{\unskip~\parencite{#1}}

\usepackage{etoolbox}

\AtBeginEnvironment{tabular}{\footnotesize}
\AtBeginEnvironment{tabularx}{\footnotesize}
\usepackage[font=small,labelfont=bf,labelsep=period,
            justification=justified,format=plain,singlelinecheck=false]{caption}
\title{Biomechanically Grounded Prediction of Selective Motor Control and Clinical Outcomes from Infant Video}

\author{
Divya Joshi\,\orcidlink{0000-0001-6397-0037}\,$^{1,2}$,
Ruize Yang\,\orcidlink{0009-0006-8353-165X}\,$^{1,3}$,
Colleen Peyton\,\orcidlink{0000-0003-4753-7246}\,$^{2,4}$,
R. James Cotton\,\orcidlink{0000-0001-5714-1400}\,$^{1,5}$}

\affiliations{
$^{1}$Center for Bionic Medicine, Shirley Ryan AbilityLab, Chicago, IL \\
$^{2}$Dept. of Physical Therapy \& Human Movement Science, Northwestern University, Chicago, IL \\
$^{3}$Dept. of Neuroscience, Northwestern University, Evanston, IL \\
$^{4}$Dept. of Pediatrics, Northwestern University, Chicago, IL \\
$^{5}$Dept. of Physical Medicine \& Rehabilitation, Northwestern University, Chicago, IL}

\paperdate{November 2026}

\correspondence{divyajoshi@northwestern.edu}

\paperabstract{\theabstract}

\begin{document}

\maketitle
\thispagestyle{empty}
\pagestyle{empty}

%%%%%%%%%%%%%%%%%%%%%%%%%%%%%%%%%%%%%%%%%%%%%%%%%%%%%%%%%%%%%%%%%%%%%%%%%%%%%%%%
\section{INTRODUCTION}

Cerebral palsy (CP) is the most common motor disability of childhood, and early identification plays a critical role in early intervention while the developing nervous system is highly plastic \cite{Novak2017,Herskind2015EarlyPalsy}. Clinical assessment tools focus on spontaneous infant movement, particularly the quality, variability, and selectivity of movements that emerge during early motor development. Two notable tools are the Prechtl General Movement Assessment (GMA), which evaluates the global quality of spontaneous movements during the writhing and fidgety movement phases \cite{Bosanquet2013AChildren,Einspeiler2005}, and the Baby Observational Selective Control AppRaisal (BabyOSCAR) \cite{Peyton2024BabyYears,SukalMoulton2024BabyPalsy,Barbosa2024BabyPerformance,peyton2026selective} , which evaluates movement at the level of individual joint axes. 

Both instruments require trained raters and substantial scoring time, limiting their scalability and availability across clinical and research settings. The GMA provides a global assessment of movement quality as a singular classification, whereas the BabyOSCAR evaluates the capacity for independent joint control of 32 joints, offering finer resolution into independent or non-independent joint movement patterns (synergy, mirroring, sinuous motion). Automating these assessments could enable efficient, scalable screening for infants with higher probability of CP, while preserving clinically meaningful joint-level information.

Recent work has used computer vision and pose estimation to predict clinical outcomes from infant video \cite{Chambers2020ComputerRisk,GamaAutomaticMethods,Segado2026AVideos,Alfano2026TheInfants}. However, all existing approaches use sparse two-dimensional keypoint representations and hand-engineered kinematic descriptors \cite{joshi2026_mmc_infants}, which do not directly encode depth or the full three-dimensional configuration of the body. Consequently, kinematics derived from these points are estimates of image-plane motion rather than measurements of a biomechanical pose. This limitation is particularly important for the BabyOSCAR, whose clinical interpretation depends on the relative three-dimensional behavior of individual joints and on whether movement occurs independently or as part of a broader pattern. In addition, feature-engineering approaches constrain the analysis to motion descriptors hand-selected in advance, potentially preventing the model from discovering clinically relevant patterns in the underlying movement trajectories. Finally, prior automated studies have focused primarily on GMA or CP-related outcomes; to our knowledge, automation of the BabyOSCAR has not previously been attempted.

In this work, we address these limitations by using a biomechanical pose representation derived from monocular video. Our infant-adapted pipeline combines our previously established cross-model distillation \cite{cotton2026crossmodeldistillation} with a biomechanical head trained through inverse-kinematics fitting \cite{cotton2026b3dbody} to estimate anatomically defined joint motion without requiring paired image-to-biomechanics labels. We term this combined framework Baby Biomechanical 3D Body (BB3D). This representation provides a common basis for validating pose estimates and predicting granular clinical movement scores. We investigate whether this biomechanical representation can support both reliable pose estimation and clinically relevant movement assessment.

The study has two objectives. First, we validate monocular biomechanical estimates against recordings acquired with our established multi-view markerless motion capture (MMMC) system \cite{Cotton2023MarkerlessPipeline} in a prospective cohort of 17 infants (33 sessions, 345 recordings total), following a previously described acquisition and preprocessing protocol \cite{joshi2026_mmc_infants}. The multi-camera reconstruction provides a reference frame for evaluating the accuracy of the marker locations and joint trajectories inferred from a single camera. Second, we train a spatio-temporal transformer \cite{motionbert2022} on the biomechanical trajectories estimated using single-camera recordings from a retrospective cohort (574 infants, 1,284 recordings) to predict clinically meaningful outcomes, including the BabyOSCAR total and per-item scores, CP diagnosis, functional ability, and body distribution. We evaluate various model architecture formulations and compare the resulting performance with existing pose-based and human scoring approaches.

The main contributions of this work are:

\begin{enumerate}

    \item \textbf{Multi-camera validation of monocular biomechanical pose estimation in infants.} We compare an infant-adapted monocular pose pipeline (BB3D) against a calibrated multi-camera reference and quantify the accuracy of its estimated joint centers and marker sites.

    \item \textbf{Automated prediction of joint-level selective motor control.} We develop a spatio-temporal transformer that predicts the 32 per-joint BabyOSCAR items from the biomechanical joint-angle trajectories BB3D estimates in single-camera recordings, and compare its scores with expert ratings in agreement and in dimensionality.

    \item \textbf{Clinical outcomes from predicted scores.} We use scores derived from the predicted items to classify CP diagnosis, functional ability and body distribution, and compare their performance with a replicated 2D-keypoint classifier and with the same scores computed from the expert's ratings.
\end{enumerate}

Together, these contributions test whether biomechanically grounded movement representations can make infant clinical assessment more scalable while retaining the joint-level information needed to characterize selective motor control.

\section{METHODS}

\subsection{Participants}

This study included infants from seven datasets collected across clinical, laboratory, and home-based settings. The cohort included both preterm and full-term infants, recorded at 28--64 weeks post-menstrual age (PMA). Several datasets included longitudinal recordings, so the same infant could contribute videos at different ages. The multi-camera validation cohort included 17 infants from 33 recording sessions, with 345 recordings evaluated. The single-camera clinical prediction cohort included 574 infants and 1,284 recordings: 1,163 recordings from 505 infants for pretraining, of which 550 BabyOSCAR-scored recordings from 350 infants were used for fine-tuning, and a held-out test set of 121 recordings from 69 infants. Demographic and clinical characteristics of each set, including CP diagnosis, Gross Motor Function Classification System (GMFCS) level, CP body distribution, sex, birth type, and PMA at recording, are summarized in Table~\ref{tab:tbl1}.

Cross-validation folds were grouped by infant, so no infant contributed to both training and validation folds. Twenty-two infants who also had training recordings were excluded from the test set, so no test infant contributed recordings to pretraining or fine-tuning. Of the remaining 69 test infants, 68 had recordings at 50--60 weeks PMA with complete expert BabyOSCAR scores and CP diagnosis, GMFCS, and body distribution information, and were included in the clinical-outcome analyses (Fig.~\ref{fig4}).

All procedures involving human participants were approved by the relevant institutional review board or research ethics committee at each contributing site (Northwestern University, Lurie Children's Hospital, University of Chicago Comer Children's Hospital, and Shirley Ryan AbilityLab). Informed consent was obtained from parents or legal guardians or waived by the relevant institutional review board, as appropriate.

\begin{table*}
    \caption{\textbf{Characteristics of the training and test cohorts.} Demographic and clinical characteristics of the cohorts used for transformer pretraining and BabyOSCAR fine-tuning. Training infants were evaluated out-of-fold during cross-validation, whereas test infants were held out. Age at assessment denotes post-menstrual age (PMA) at recording, in weeks. Categorical variables were compared using $\chi^2$ tests and age using Welch’s \textit{t}-test. SMD denotes the standardized mean difference between the fine-tuning training and test groups; values below 0.1 indicate good balance \cite{Yang2012}}
    \centering
    \begin{tabularx}{\textwidth}{@{}Xcccccc@{}}
        \toprule
        \textbf{Characteristic} & \textbf{Pretraining} & \textbf{Fine-tuning} & \textbf{Test} & \textbf{Statistic} & \textbf{$p$} & \textbf{SMD}\\
        \midrule
        Infants, $n$ & 505 & 350 & 69 &  &  & \\
        Recordings, $n$ & 1163 & 550 & 121 &  &  & \\
        \addlinespace
        CP diagnosis, $n$ &  &  &  & $\chi^2=0.59$ & 0.898 & 0.10\\
        \quad CP          & 134 & 128 & 22 &  &  & \\
        \quad Likely CP   & 17  & 17  & 4  &  &  & \\
        \quad NCP         & 223 & 152 & 32 &  &  & \\
        \quad Unlikely CP & 53  & 53  & 11 &  &  & \\
        \addlinespace
        GMFCS level, among CP, $n$ &  &  &  & Fisher's exact & 0.186 & 0.52\\
        \quad I--II   & 36 & 34 & 7  &  &  & \\
        \quad III--V  & 44 & 41 & 3  &  &  & \\
        \quad Unknown & 54 & 53 & 12 &  &  & \\
        \addlinespace
        CP distribution, $n$ &  &  &  & $\chi^2=0.12$ & 0.726 & 0.14\\
        \quad Unilateral & 48 & 44 & 9  &  &  & \\
        \quad Bilateral  & 78 & 78 & 12 &  &  & \\
        \quad Unknown    & 8  & 6  & 1  &  &  & \\
        \addlinespace
        Sex, $n$ &  &  &  & $\chi^2=1.71$ & 0.191 & 0.19\\
        \quad Female  & 184 & 158 & 25 &  &  & \\
        \quad Male    & 244 & 188 & 44 &  &  & \\
        \quad Unknown & 77  & 4   & 0  &  &  & \\
        \addlinespace
        Birth type, $n$ &  &  &  & $\chi^2=0.08$ & 0.775 & 0.06\\
        \quad Preterm & 277 & 224 & 46 &  &  & \\
        \quad Term    & 155 & 126 & 23 &  &  & \\
        \quad Unknown & 73  & 0   & 0  &  &  & \\
        \addlinespace
        Age at assessment, weeks PMA, \\mean $\pm$ SD & $52.4 \pm 4.0$ & $52.2 \pm 3.8$ & $52.2 \pm 4.1$ & $t=-0.11$ & 0.912 & 0.01\\
        \bottomrule
    \end{tabularx}

\label{tab:tbl1}
\end{table*}

\subsection{Video Acquisition and Preprocessing}

\subsubsection{Multi-Camera Recordings}

Multi-camera recordings were acquired with our MMMC system, which captures spontaneous infant movement from eight viewpoints \cite{joshi2026_mmc_infants,Cotton2023MarkerlessPipeline}. Infants lay supine in a calm, alert state; each session comprised 8--10 trials of 50--60 seconds. Calibration, synchronization and tracking followed our previous procedures \cite{joshi2026_mmc_infants,Cotton2023MarkerlessPipeline}.

Two-dimensional keypoints were detected in every view with Sapiens2 \cite{khirodkar2026sapiens2}, the successor to Sapiens, which performed best for dense infant keypoints in our earlier comparison \cite{joshi2026_mmc_infants,Khirodkar2024Sapiens:Humans.}. Keypoints from the seven cameras other than the top-down camera were robustly triangulated \cite{Roy2022OnEstimation,joshi2026_mmc_infants}, treating points with confidence below 0.5 as unobserved, so that the reference was independent of the evaluated view. The triangulated keypoints were reprojected into the top-down camera as the reference for its monocular estimates. BB3D \cite{cotton2026b3dbody,cotton2026crossmodeldistillation} was applied to each top-down recording to estimate body pose, segment scales and marker offsets, which forward kinematics converted to 3D joint centers and marker sites.

\subsubsection{Single-Camera Recordings}

The single-camera cohort comprised 1--5-minute videos from laboratory, clinical, neonatal intensive care unit (NICU) and home settings, so camera placement, image quality, background, lighting and clothing varied. The infant was tracked with the Segment Anything Model 3 (SAM 3) prompted with ``baby'' \cite{sam3}, initialized on the highest-scoring detection, and cropped for pose estimation; this avoided tracking caregivers or other people in the frame. The crops were processed by the infant-adapted BB3D pipeline \cite{cotton2026crossmodeldistillation,cotton2026b3dbody}, with Sapiens providing 2D keypoints \cite{Khirodkar2024Sapiens:Humans.} and BB3D the 3D body configuration, converted to joint-angle trajectories that served as input to the clinical prediction models.

\subsection{Validation of the Biomechanical Pose Estimates}

BB3D estimates from the top-down camera, whose viewpoint most closely approximates the overhead view used in clinical and home recordings, were compared with the triangulated reference at two levels: (1) anatomical joint centers (shoulder, elbow, wrist, hip, knee and ankle, bilaterally; 12 points), compared with the corresponding Sapiens2 joint keypoints; and (2) marker sites (34 sites across the body, face and fingertips), each paired with the corresponding Sapiens2 keypoint (Fig. \ref{fig1}A). Because marker sites and image keypoints are defined differently (skin over bony landmarks versus learned image landmarks), a pair denotes the closest equivalent point rather than an identical one.

\subsubsection{Procrustes-aligned error calculation}

Because a single view does not determine absolute body size, predicted points were aligned to the reference in each frame by a Procrustes-alignment transform (rotation, translation and uniform scale), fitted separately for each level on the pairs observed in that frame (with a minimum of six needed for evaluation). The error of a point is defined as its Euclidean distance from the reference after alignment, averaged over the frames of a recording. To separate consistent from variable disagreement, residuals are expressed in a trunk-fixed frame defined by the reference hips and shoulders; the \emph{systematic offset} is the magnitude of the mean residual vector over a recording, and the \emph{spread} is the root-mean-square residual about it. Errors are summarized as medians and inter-quartile ranges across recordings, over all points and per anatomical location (left and right pooled).

\subsubsection{Marker site-offset calibration}

Sites were evaluated at two stages computed from the same BB3D pose and segment scales. At the \emph{standard} stage, marker sites were placed at their default positions in the model's marker configuration and joint centers at their segment origins, discarding BB3D's predicted marker offsets. At the \emph{calibrated} stage, each site was displaced by one static offset in the unscaled frame of its segment, shared across all frames and recordings. Offsets were fitted to the triangulated Sapiens2 keypoints by alternating per-frame Procrustes alignment with a closed-form least-squares update of the offsets (Huber loss, 30~mm; weak ridge penalty toward the standard positions), separately for marker sites and joint centers, and leave-one-participant-out. Because the pose was unchanged, any change in error reflects site placement rather than the estimated pose. For reference, we also report the BB3D Body output with its own predicted offsets. Changes between stages were tested with a Wilcoxon signed-rank test on participant-level medians.

\subsection{Prediction of Clinical Scores and Outcomes}

Clinical prediction used the single-camera recordings and compared two models built on different representations of movement: a baseline classifier using 2D keypoints estimated with Sapiens \cite{Khirodkar2024Sapiens:Humans.}, and our transformer using biomechanical joint-angle (\verb|qpos|) trajectories estimated with BB3D Body \cite{cotton2026b3dbody,cotton2026crossmodeldistillation}. We predicted the 32 binary BabyOSCAR items and the aggregate scores derived from them, as well as clinical outcomes: CP diagnosis, functional ability (GMFCS level), and CP body distribution.

\subsubsection{Data splitting}

All models shared one participant-level 80/20 train/test split, stratified on CP diagnosis, so every recording of an infant fell on one side. The training side was divided into five cross-validation folds, grouped by participant and by duplicate-recording group. The folds were used to (1) pretrain one transformer model per fold without that fold's participants; (2) fine-tune one model per fold on the other four folds, using the held-out fold to select the stopping epoch; (3) obtain out-of-fold predictions for every training-side recording from the model that did not train on it; (4) predict the test set as the average of the five fold models; and (5) cross-fit the regression readings of clinical outcomes. The test set was used only for final evaluation.

\subsubsection{Baseline classifier}

To compare with prior work, we computed the 38 movement features reported by Segado \textit{et al.} \cite{Segado2026AVideos} using the 2D Sapiens keypoints. This feature set was classified with the \texttt{auto\_sklearn2} AutoML package \cite{auto-sklearn2} (one-hour search), trained on the training side to separate infants with and without CP, and scored once on the test side. Its per-recording outputs were averaged to the infant level in the same way as the transformer's, so both were evaluated on the same test infants.

\subsubsection{BabyOSCAR prediction with a pretrained spatio-temporal transformer}

We used 44 BB3D joint-angle (\verb|qpos|) channels at 30~Hz, excluding root, finger, metatarsophalangeal, and knee-rotation channels. The transformer was pretrained as a masked autoencoder on randomly sampled 10-second windows. Each window was divided into 30 patches of 10 frames, producing $1{,}320$ channel--patch tokens. The two-layer, 256-dimensional, four-head encoder used MotionBERT's factorized spatio-temporal attention \cite{motionbert2022}. Seventy percent of channels were masked for the full window, and a one-layer decoder reconstructed them. Pretraining ran for 200 epochs using all eligible processed recordings, excluding test participants and the held-out fold.

For fine-tuning, the decoder was removed and each of the 550 scored training recordings was represented by six evenly spaced 10-second windows. The encoder processed each window with all channels visible. Time-averaged channel embeddings were normalized and averaged across windows, yielding one 256-dimensional embedding per channel. Each BabyOSCAR item used a learned weighted combination of these embeddings, initialized from the item's joint, neighboring, and contralateral channels; items without corresponding input channels were initialized using all channels. The item representation was combined with learned joint-type and body-side tags and passed through a shared MLP to predict the item logit. Fine-tuning updated the final encoder layer and prediction components using gradient-norm-balanced binary cross-entropy across the 32 items. We applied left--right mirroring augmentation with probability 0.5, selected checkpoints using the held-out training fold, trained three random seeds, and averaged their test-set probabilities.

\paragraph{Derived BabyOSCAR scores.} The total was the sum of the 32 item probabilities of independent joint motion including limb totals, right and left summed joint totals and a laterality index (right minus left) were derived from the items as in the manual assessment \cite{Peyton2024BabyYears,SukalMoulton2024BabyPalsy,Barbosa2024BabyPerformance,peyton2026selective}. To compare the structure of predicted and expert scores, we ran a principal component analysis on the 32-item correlation matrix of held-out recordings (out of fold and test) with all items rated, separately for expert scores and predicted probabilities. We report cumulative variance explained, the components needed for 80\% and 90\%, and the participation ratio, an effective number of dimensions computed from the eigenvalue spread (1 if one component carries all variance, 32 if it is spread evenly). Bands are participant-clustered bootstrap 95\% intervals (200 resamples).

\subsubsection{Clinical outcomes from predicted BabyOSCAR scores}

Outcomes were read at the infant level from recordings at 50--60 weeks post-menstrual age (PMA), averaged within each session and then across sessions. Readings were fixed scores (the negative total, or the asymmetry $\sum|R-L|$) or regularized regressions on six summaries or all 32 items, cross-fitted over the folds. Tasks were CP versus no CP (NCP), and, among infants with CP, GMFCS III--V versus I--II and unilateral versus bilateral CP. Additionally, each CP subgroup (GMFCS III--V, GMFCS I--II, unilateral, bilateral) was compared with the NCP group, using the negative total or the asymmetry. The same readings were taken from the expert's item scores on the same infants and compared with the model's using a paired, infant-level bootstrap, separately for the test set and the out-of-fold training side.

\subsubsection{Model experiments}

Each experiment changed one aspect of the base model, keeping its folds, encoders and seeds:
\begin{itemize}
    \item adding CP diagnosis and GMA rating heads;
    \item adding heads that predict the total, limb totals, side totals and laterality directly;
    \item training only on recordings at or after 40 weeks PMA, evaluated on the same $\geq$40-week recordings;
    \item replacing the shared item network with one network per item.
\end{itemize}

\subsubsection{Evaluation}

Because recruitment datasets differed in referral pathway and outcome prevalence, headline metrics were computed within dataset, with pooled values alongside. Items were scored by ROC-AUC within each dataset of at least 20 recordings, weighted by discordant pairs and averaged over the 28 items whose joints the channels measure. The total and derived scores were scored by the mean within-dataset Spearman correlation, with mean absolute error and calibration slope. Out-of-fold predictions are mildly optimistic, because the same fold selected the stopping epoch; test predictions averaged the five fold models. Model experiments were compared with the base model on the same recordings, using participant-clustered bootstrap confidence intervals.

\section{RESULTS}

\subsection{Validation of the Biomechanical Pose Estimates}

\begin{figure*}[!tbp]
    \centering
    \includegraphics[width=\textwidth]{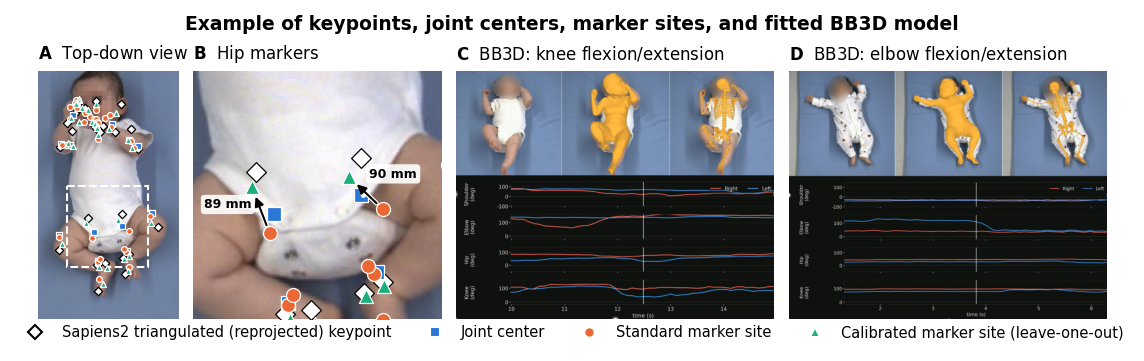}
    \caption{\textbf{Example of keypoints, joint centers, marker sites and the fitted BB3D model on one top-down camera frame.} \textbf{(A)} Sapiens2 keypoints triangulated from the seven non-top-down cameras and reprojected into the top-down view (white diamonds), with the corresponding BB3D joint centers (blue squares), standard marker sites (orange circles) and calibrated marker sites (green triangles). The dashed box marks the region enlarged in (B). \textbf{(B)} Hips close-up. Arrows run from each standard hip marker site to its calibrated position, labeled with the 3D displacement in mm. \textbf{(C, D)} BB3D fit to the video (frame, mesh overlay, skeleton overlay) with the resulting joint-angle time series for the right (red) and left (blue) side. (C) highlights knee and (D) elbow flexion/extension.}
    \label{fig1}
\end{figure*}

\begin{figure*}[!tbp]
    \centering
    \includegraphics[width=\textwidth]{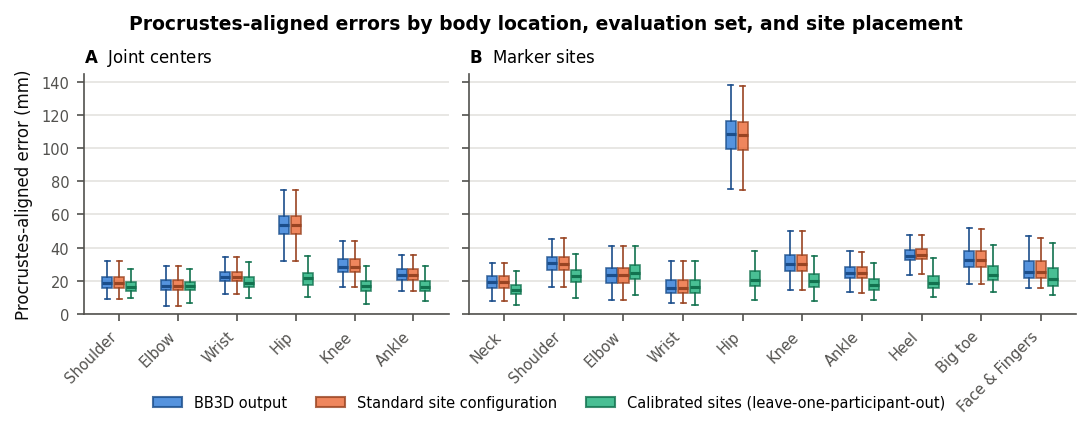}
    \caption{\textbf{Procrustes-aligned errors of fitted BB3D by body location, evaluation set, and site placement.}  Errors are measured against Sapiens2 keypoints triangulated from the seven other cameras, after per-frame similarity alignment (rotation, translation and uniform scale), for \textbf{(A)} the 12 joint centers and \textbf{(B)} the 34 marker sites. Each box summarizes metrics over 345 recordings from 17 participants: the mean error over that body location's sites, with left and right pooled. Nose, eyes, ears, thumb and fingertips are pooled as "Face \& Fingers". Colors compare the BB3D output (blue), the standard site configuration (orange) and calibrated sites (green). Calibration offsets were fitted on all other participants (leave-one-participant-out). Each set is aligned per frame on its own points. Boxes show the median and interquartile range; whiskers extend to 1.5× the interquartile range; outliers are not shown.}
    \label{fig2}
\end{figure*}

We analyzed 345 recordings from 17 participants; the held-out camera's Sapiens2 detections had a median reprojection error of 5.4 pixels (IQR 2.1). With the standard model configuration, the median Procrustes-aligned error was 26.9~mm for joint centers and 30.5~mm for marker sites, mostly systematic (offsets 24.8 and 25.9~mm vs. spreads 10.9 and 16.5~mm) and largest at the hips (Fig. \ref{fig1}A, B, and \ref{fig2}). Leave-one-participant-out calibration reduced these to 17.5 and 20.4~mm, improving all 17 participants (Wilcoxon $p<0.0001$; Figs.~\ref{fig1}B and \ref{fig2}), by correcting the systematic offset (13.6 and 14.5~mm) while leaving the spread unchanged. Reduction was largest at the hips (Figs.~\ref{fig1}B and \ref{fig2}). Held-out offsets performed almost identically to offsets fitted on all participants (17.5 vs. 17.3~mm and 20.4 vs. 20.1~mm).

\subsection{Automated BabyOSCAR scoring}

\begin{figure*}[!tbp]
    \centering
    \includegraphics[width=\textwidth]{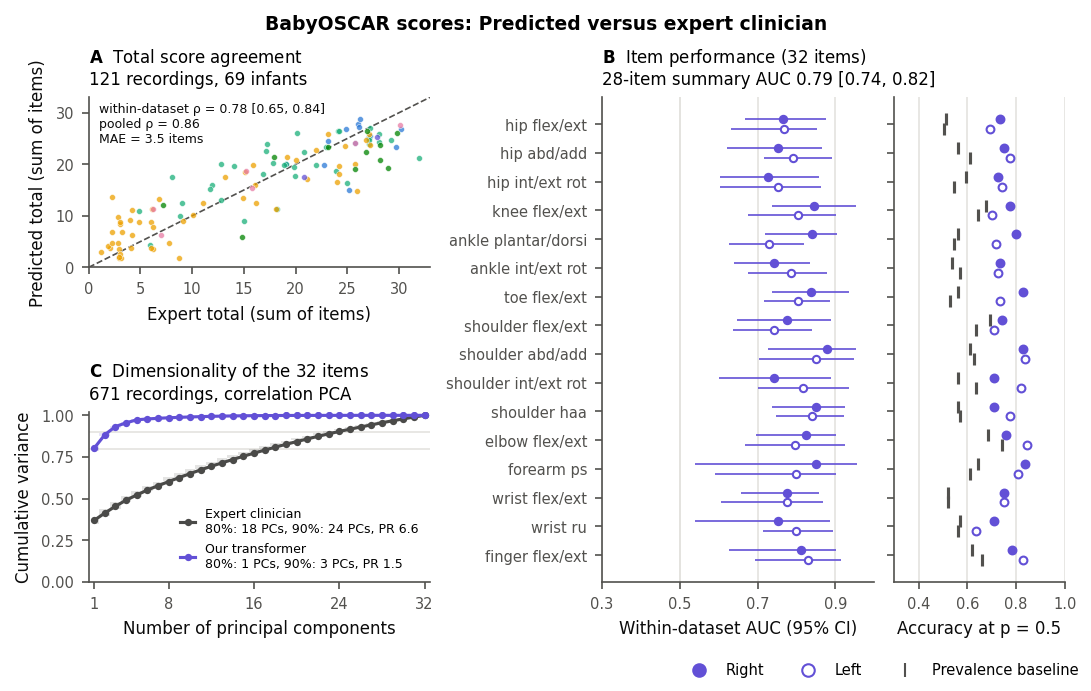}
    \caption{\textbf{BabyOSCAR score predictions on monocular recordings.} \textbf{(A)} Predicted total score (sum of the 32 predicted item probabilities) against the expert total, one point per recording (121 recordings, 69 infants) colored by recruitment dataset. The dashed line marks perfect agreement. within-dataset Spearman $\rho$ with 95\% CI, the pooled $\rho$ and the mean absolute error are reported. \textbf{(B)} Per-item performance for the 16 joint degrees of freedom, right (filled) and left (open). Left: within-dataset AUC with 95\% CI. Right: accuracy at a 0.5 threshold beside the prevalence baseline (always predicting the more common class). The title gives the 28-item summary AUC.  \textbf{(C)} Cumulative variance explained by principal components of the 32 item scores: expert labels (gray) against predicted probabilities (purple). This panel pools out-of-fold and test recordings (671 recordings), because dimensionality is not a held-out score. Shaded bands are 95\% participant-clustered bootstrap intervals. The legend gives the components needed for 80\% and 90\% of the variance and the participation ratio (PR).}
    \label{fig3}
\end{figure*}

On the 121 held-out test recordings from 69 infants, the predicted BabyOSCAR total, computed as the sum of the 32 predicted item probabilities, agreed closely with the expert's total (within-dataset Spearman $\rho$ = 0.78, 95\% CI [0.65-0.84]; pooled $\rho$ = 0.86), with a mean absolute error of 3.5 items (Fig. \ref{fig3}A). Agreement extended to the individual items. Across the 28 items whose joint is tracked by BB3D, the mean within-site AUC was 0.79 (0.74-0.82) (Fig. \ref{fig3}B). Item AUCs ranged from 0.73 for right hip internal/external rotation to 0.88 for right shoulder abduction/adduction, and all 32 items exceeded the accuracy of always predicting the most common score at a 0.5 threshold.

The predicted item scores, however, had a lower dimensionality than the expert's. Principal component analysis showed that 18 components explained 80\% of the variance in the expert's 32 item scores (participation ratio 6.6), compared with just 1 component for the predicted scores (participation ratio 1.5) (Fig. \ref{fig3}C).

\subsection{Clinical outcomes from predicted BabyOSCAR scores}

Next, we investigated whether the predicted item scores carry the same clinical outcome information as the expert's. Each infant was represented by a single 32-item vector, the mean of their predicted item probabilities over sessions at 50--60 weeks PMA. This window corresponds to the fidgety-movement period, in which GMA is assessed \cite{Einspeiler2005} and in which the BabyOSCAR has been validated for outcome prediction \cite{Barbosa2024BabyPerformance,SukalMoulton2024BabyPalsy,Peyton2024BabyYears,peyton2026selective}. To avoid fitting a model to the small clinical groups, each outcome was read from a fixed summary of the items, as the BabyOSCAR is read clinically. The total score was used for CP diagnosis and classification of functional ability, with lower totals indicating greater need for mobility support. Asymmetry, the sum over the 16 movements of the absolute right–left difference, was used for distinguishing body distribution types (unilateral vs. bilateral) in the CP versus NCP groups, as well as classifying body distribution within the CP subset. The expert's item scores were summarized identically on the same infants, providing a reference for the clinical information contained in the BabyOSCAR itself.

\begin{figure*}[!tp]
    \centering
    \includegraphics[width=\textwidth]{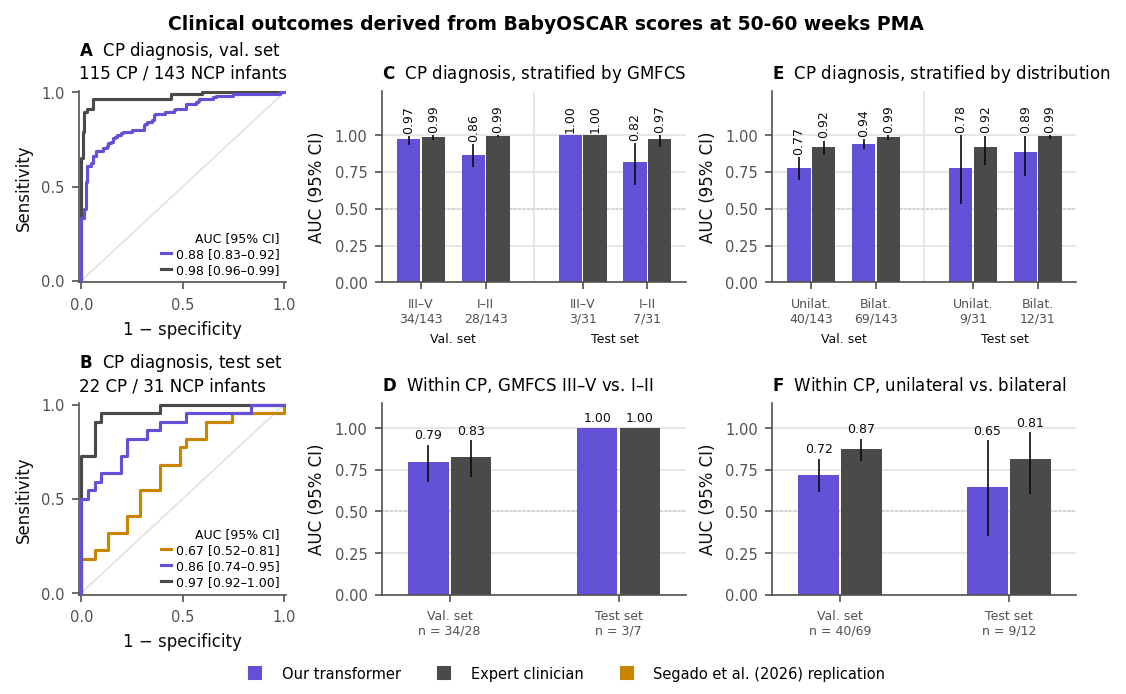}
    \caption{\textbf{Infant-level prediction of clinical outcomes at 50–60 weeks PMA.}
    These are pre-registered fixed readings of the BabyOSCAR items; no model is fitted to the outcome. Both our transformer's predicted items (purple) and the expert's item scores (gray) are read the same way, on the same infants:
(1) the negative total score for CP diagnosis and GMFCS level, and (2) the left–right asymmetry $\Sigma|R-L|$ for laterality. Gold is our implementation of Segado \textit{et al.} (2026): 38 movement features with a classifier trained on CP vs. NCP, scored on the test set only. \textbf{(A, B)} ROC curves for CP vs. NCP  on the validation set (out of fold: each infant scored by the fold model that did not train on it) and the held-out test set; the dotted line marks chance (0.5). \textbf{(C)} CP diagnosis, stratified by GMFCS level (GMFCS III–V vs. NCP, GMFCS I-II vs. NCP).  \textbf{(D)} GMFCS level, among CP group. \textbf{(E)} CP diagnosis, stratified by body distribution (unilateral vs. NCP, bilateral vs. NCP). \textbf{(F)} Body distribution, among CP group. Bars give the AUC with 95\% CIs from an infant-clustered bootstrap (2,000 resamples), and the light gray line marks chance (0.5). Each infant is counted once, and infants with recordings on both sides of the train/test split are excluded.}
    \label{fig4}
\end{figure*}

Each outcome was evaluated on two sets of infants (Fig.~\ref{fig4}). The primary analysis used the 68 test-set infants, which no model saw during training or model selection. Because the test set contains few infants with CP, we also report the 324 validation-set infants, each predicted by the cross-validation model that had not been trained on them. The fine-tuning epoch for each cross-validation model was chosen using the item loss on its validation fold, so training-set estimates may be slightly optimistic; the test-set estimates are not affected by this selection.

On the test set, the predicted total separated infants later diagnosed with CP (n=22) from infants without CP (n=31) with an AUC of 0.86 (95\% CI 0.74–0.95), compared with 0.97 (0.92–1.00) for the expert's total and 0.67 (0.52–0.81) for our implementation of the 38-feature classifier of Segado \textit{et al.} \cite{Segado2026AVideos}, trained directly on CP versus NCP and evaluated on the same test set of infants (Fig.~\ref{fig4}B). On the validation set (115 CP, 143 NCP), the AUCs were 0.88 (0.83–0.92) for the predicted total and 0.98 (0.96–0.99) for the expert's (Fig.~\ref{fig4}A).

Classification of CP diagnosis using stratification based on functional ability (GMFCS level) and body distribution showed that separation from the NCP group was strongest for those with GMFCS III-V and for bilateral CP (Fig.~\ref{fig4}C, E). On the validation set, the predicted total reached an AUC of 0.97 for GMFCS III–V versus NCP and 0.86 for GMFCS I–II versus NCP (expert 0.99 for both); predicted asymmetry separated unilateral CP versus NCP with an AUC of 0.77 (expert 0.92), and 0.94 for bilateral CP from NCP (expert 0.99). The test set showed the same pattern (GMFCS III–V 1.00, I–II 0.82,  unilateral 0.78, and bilateral 0.89; expert 1.00, 0.97, 0.99 and 0.92), on 3, 7, 9 and 12 infants with CP, respectively.

Among infants with CP, the predicted and expert totals both distinguished GMFCS III–V from GMFCS I–II with an AUC of 1.00 on the test set (3 vs.\ 7 infants), and with AUCs of 0.79 and 0.83 on the validation set (34 vs.\ 28) (Fig.~\ref{fig4}D). Predicted asymmetry distinguished unilateral from bilateral CP with an AUC of 0.65 (0.35–0.93) on the test set (9 vs.\ 12 infants) and 0.72 on the validation set (40 vs.\ 69), compared with 0.81 and 0.87 for the expert's scores (Fig.~\ref{fig4}F).

\subsection{Model Experiments}

Finally, to test which design choices contributed to performance, we compared four variants with the reference model on the same recordings and infants: (1) additional CP and GMA heads, (2) additional heads for the total, limb, side and laterality scores, (3) training only on recordings at or beyond 40 weeks PMA, and (4) a separate classifier for each item. None of these changes significantly altered performance for any of the metrics; for every metric, the 95\% confidence interval of the difference from the reference included zero. We therefore retained the simpler reference design.

\section{DISCUSSION}

This study shows that three-dimensional biomechanical information estimated from monocular video can predict joint-level selective motor control and clinical outcomes in infants. Unlike previous approaches based on image-plane keypoints \cite{Chambers2020ComputerRisk,GamaAutomaticMethods,Segado2026AVideos,Alfano2026TheInfants}, our representation describes how each joint moves in three dimensions, the level at which the BabyOSCAR is defined.

Compared with a multi-camera reference, BB3D estimates from a single camera agreed to within approximately 2~cm after calibration of marker-site placement (17.5~mm for joint centers, 20.4~mm for marker sites). Calibration removed mostly systematic error, left frame-to-frame spread unchanged and transferred to held-out infants, indicating that residual error largely reflects differences in landmark definition rather than pose.

The predicted BabyOSCAR agreed well with expert scoring (within-dataset $\rho = 0.78$ for the total; mean item AUC 0.79), and, although trained only on BabyOSCAR items, separated CP from NCP (AUC 0.86). Pooled agreement exceeded within-dataset agreement ($\rho$ = 0.86 vs.\ 0.78), but this need not reflect site confounding: the transformer sees only scale-normalized joint-angle trajectories, so image-level shortcuts are unlikely, and each dataset source spans a narrower range of ages and impairment. Within-dataset agreement is thus a conservative lower bound, and pooled agreement reflects sensitivity across the full developmental spectrum.

A replication of the 2D feature classifier of Segado \textit{et al.} \cite{Segado2026AVideos}, without depth or temporal structure, achieved an AUC of 0.67 for CP versus NCP on the same test infants, compared with 0.86 for the transformer. The expert nevertheless remained more accurate, and the structure of the scores suggests why. The expert's item scores had a participation ratio more than four times that of the predicted scores (6.6 vs. 1.5 effective dimensions). The model compresses the BabyOSCAR onto a few shared axes, dominated by global movement quality, whereas the clinician scores each joint using information the model does not receive: whether a movement is isolated or part of a mirroring or synergy pattern, its timing, and the context of the whole observation \cite{Peyton2024BabyYears,SukalMoulton2024BabyPalsy,Barbosa2024BabyPerformance,peyton2026selective}. Part of the expert's higher dimensionality may also be item-level rating noise, so the ratio bounds, rather than measures, what the model misses.

This difference appears in the stratified results. In distinguishing between CP, GMFCS III--V versus NCP groups, the transformer performed nearly as well as the expert (AUC 0.97 vs. 0.99), as infants who have higher GMFCS levels have less independent motion expressed across many joints, captured by a global axis. The gap widened for infants with CP with GMFCS I-II and for distinctions that depend on specific joints or sides, with higher amounts of independent joint motion and laterality within CP. Automated scoring could therefore already support scalable screening for infants with GMFCS levels III-V, while infants with GMFCS I-II and unilateral presentations still benefit from expert assessment.

\subsection{Limitations}

The model lacks event detection and timing and does not label the movement patterns, such as mirroring or synergies, that explain its scores. Source differences in age and impairment mean that pooled and within-dataset metrics bound performance rather than measuring it exactly. GMFCS level and body distribution information were often missing, so within-CP and subgroup comparisons rest on few infants and wide confidence intervals. Monocular accuracy was validated only from a laboratory top-down camera, not under clinical and home conditions or in infants younger than our validation set, such as those in the NICU.

\subsection{Future Directions}

Future models could ingest clinicians' notes on movement patterns and event timing for each joint, or detect mirroring and synergies explicitly, to approach the dimensionality of expert scoring. Our datasets also include birth history information, clinical labels and longitudinal recordings, which could support causal models of infants' motor trajectories despite missing labels.

In summary, BB3D provides monocular three-dimensional estimates of infant movement within approximately 2~cm. Temporal models built on these estimates outperform image-plane approaches in predicting selective motor control and clinical outcomes, and approach expert performance in identifying infants with CP GMFCS III-V, a step toward scalable, joint-level infant motor assessment.

%%%%%%%%%%%%%%%%%%%%%%%%%%%%%%%%%%%%%%%%%%%%%%%%%%%%%%%%%%%%%%%%%%%%%%%%%%%%%%%%

\section*{ACKNOWLEDGMENT}

We would like to to thank Imani Mann and Grace Hoo for their assistance with participant recruitment and scheduling, Shawana Anarwala and Kayan Abdou for their assistance with experimental setup and data collection, and Kunal Shah for his assistance with infrastructure. Research reported in this publication was supported by the Eunice Kennedy Shriver National Institute Of Child Health \& Human Development of the National Institutes of Health under Award Number R21HD117074, and the National Center for Advancing Translational Sciences of the National Institutes of Health under Award Number T32TR005124. The content is solely the responsibility of the authors and does not necessarily represent the official views of the National Institutes of Health.

%%%%%%%%%%%%%%%%%%%%%%%%%%%%%%%%%%%%%%%%%%%%%%%%%%%%%%%%%%%%%%%%%%%%%%%%%%%%%%%%

%% =============================================================
%% Bibliography (IEEE uses bibtex/IEEEtran; ISR uses biblatex)
%% =============================================================

{\small
\printbibliography
}

\end{document}